\pdfoutput=1

\documentclass[11pt]{article}
 
\usepackage[final]{EMNLP2022}
\usepackage{times}
\usepackage{latexsym}
\usepackage[T1]{fontenc}
\usepackage[utf8]{inputenc}
\usepackage{microtype}
\usepackage{inconsolata}

\usepackage{lineno}

\usepackage[ruled]{algorithm2e}
\usepackage{booktabs}
\usepackage{times}
\usepackage{graphicx}
\usepackage{multirow}
\usepackage{multicol}
\usepackage{hyperref}
\usepackage{makecell}

\usepackage[flushleft]{threeparttable}
\usepackage{amsmath,amssymb,amsfonts}
\usepackage{scalerel,xparse}

\makeatletter
\def\thanks#1{\protected@xdef\@thanks{\@thanks
        \protect\footnotetext{#1}}}
\makeatother

\title{\textsc{When FLUE Meets FLANG}: Benchmarks and Large Pre-trained Language Model for Financial Domain}

\author{\hypersetup{linkcolor=black} Raj Sanjay Shah$^{\dagger}$\;, Kunal Chawla$^{\dagger * }$\;, Dheeraj Eidnani$^{\dagger *}$\;, Agam Shah$^{\dagger *}$\;, Wendi Du$^{\dagger}$\\ \bf \hypersetup{linkcolor=black} Sudheer Chava$^{\dagger}$\;, Natraj Raman$^{\clubsuit}$\;, Charese Smiley$^{\clubsuit}$\;, Jiaao Chen$^{\dagger}$\;, Diyi Yang$^{\heartsuit}$\\
$^{\dagger}$ Georgia Institute of Technology\\
$^{\clubsuit}$ JPMorgan AI Research\\
$^{\heartsuit}$ Stanford University
\thanks{Email IDs of the authors: \textcolor{darkblue}{{\{\href{mailto:rajsanjayshah@gatech.edu}{rajsanjayshah},
\href{mailto:kunalchawla@gatech.edu}{kunalchawla}, \href{mailto:deidnani@gatech.edu}{deidnani}, \href{mailto:ashah482@gatech.edu}{ashah482}, \href{mailto:wendi.du@gatech.edu}{wendi.du}, \href{mailto:schava6@gatech.edu}{schava6}, \href{mailto:jchen896@gatech.edu}{jchen896}\}@gatech.edu}}, \href{mailto:natraj.raman@jpmorgan.com}{natraj.raman@jpmorgan.com}, \href{mailto:charese.h.smiley@jpmchase.com}{charese.h.smiley@jpmchase.com}, \href{mailto:diyiy@cs.stanford.edu}{diyiy@cs.stanford.edu}
    }\thanks{* These authors contributed equally to this work}
    }

\date{}

\begin{document}
\maketitle

\begin{abstract}
Pre-trained language models have shown impressive performance on a variety of tasks and domains. Previous research on financial language models usually employs a generic training scheme to train standard model architectures, without completely leveraging the richness of the financial data. We propose a novel domain specific Financial LANGuage model (FLANG) which uses financial keywords and phrases for better masking, together with span boundary objective and in-filing objective. Additionally, the evaluation benchmarks in the field have been limited. To this end, we contribute the Financial Language Understanding Evaluation (FLUE), an open-source comprehensive suite of benchmarks for the financial domain. These include new benchmarks across 5 NLP tasks in financial domain as well as common benchmarks used in the previous research. Experiments on these benchmarks suggest that our model outperforms those in prior literature on a variety of NLP tasks. Our models, code and benchmark data are publicly available on Github and Huggingface\footnote{The website can be found at \href{https://salt-nlp.github.io/FLANG/}{https://salt-nlp.github.io/FLANG/}. All the FLANG models are available on the \href{https://huggingface.co/SALT-NLP/FLANG-BERT}{Huggingface SALT-NLP} site.}
\end{abstract}

\begin{figure*}
\centering
\includegraphics[width=\linewidth]{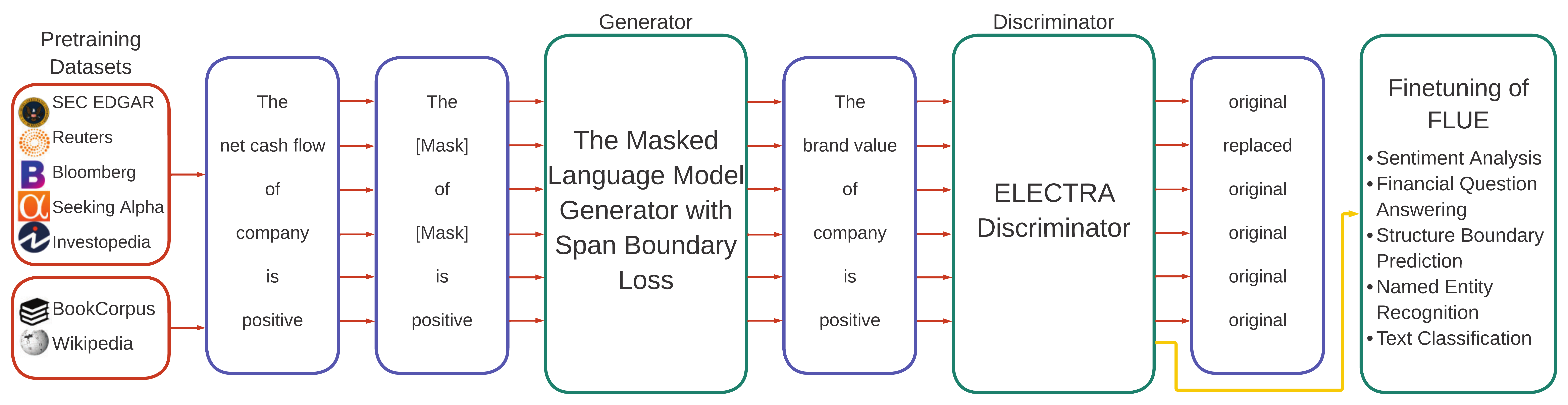}
\caption{Architecture of our model. We use finance specific datasets and general English datasets (Wikpedia and BooksCorpus) for training the model. We follow the training strategy of ELECTRA \cite{electra} with span boundary task which first predicts masked tokens using language model and then uses a discriminator to assess if a token is original or replaced. The generator and discriminator are trained end-to-end, and both words and phrases from financial vocabulary are used for masking. The final discriminator is then fine-tuned on individual tasks on our contributed benchmark suite, Financial Language Understanding Evaluation (FLUE). Note that our method is not specific to ELECTRA and can be generalized to other models.}
\vspace*{-1em}
\end{figure*}
  
\section{Introduction}
Efficient financial markets incorporate all price relevant information available to investors at that point of time. Unstructured data, such as textual data, help complement structured data traditionally used by investors. For example, in addition to quantitative data such as firm's financial performance, the tone and sentiment of firms' financial reports, earnings calls and social media posts can also influence the stock price movement \cite{bochkay2020hyperbole}. We aim to capture these textual features with the help of pre-trained deep learning models, which have shown superior performance in a variety of Natural Language Processing (NLP) tasks \cite{gpt2,bert,roberta,bart}.
However, the language used in finance and economics is likely to be different from the language of common usage. A statement like ``\textit{The crude oil prices are going up}'' has a negative sentiment for the financial markets, but it does not contain traditionally negative words such as danger, hate, fear, etc. \cite{loughran2011liability}. Therefore, it is necessary to develop a domain-specific language model training methodology that improves the performance in the downstream NLP tasks like managers' sentiment analysis and financial news classification. 

Previous research, for example, \citet{finbert, orig_finbert} have pre-trained the state-of-the-art language models like BERT \cite{bert} with financial documents, but suffer from two major limitations. First, financial domain knowledge and adaptation are not utilized in the pre-training process. We argue that the \textit{financial terminologies} play a critical role in understanding the language used in financial markets, and expect a performance improvement after incorporating the financial domain knowledge into the pre-training process. Second, the lack of different evaluation benchmarks limit the test the language models' performance in finance-related tasks.

In this work, we propose a simple yet effective language model pre-training methodology with preferential token masking and prediction of phrases.
This helps capture the fact that many financial terms are actually multi-token phrases, such as \emph{margin call} and \emph{break-even analysis}.
We contribute and make public two language models trained using this technique. Financial LANGuage Model (\texttt{FLANG-BERT})  is based on BERT-base architecture \cite{bert}, which has a relatively small memory footprint and inference time. It also enables comparison with previous works, most of which are based on BERT. We also contribute \texttt{FLANG-ELECTRA}, our best performing model, based on the ELECTRA-base architecture \cite{electra}, where we introduce a span boundary objective on the ELECTRA generator pre-training task to learn robust financial multi-word representations while masking contiguous spans of text. We show that FLANG-BERT outperforms all previous works in nearly all our benchmarks, and FLANG-ELECTRA further improves the performance giving two new state-of-the-art models. Our training methodology can be extended to other domains that would benefit from domain adaptation. 

Financial domain benchmarks are critical to evaluate the newly developed financial language models. Inspired by GLUE \cite{glue}, a set of comprehensive benchmarks across multiple NLP tasks, we construct Financial Language Understanding Evaluation (\texttt{FLUE}) benchmarks. FLUE consists of 5 financial domain tasks: financial sentiment analysis, news headline classification, name entity recognition, structure boundary detection, and question answering. We intend for this benchmark suite to be a standard for evaluation of natural language tasks in financial domain, subject to appropriate license and privacy considerations. All proposed benchmarks will be made publicly available on Github and Huggingface.

Our contributions are as follows:
\begin{itemize}\setlength\itemsep{0em}
    \item We propose masking finance-specific words and phrase masking for pre-training language model, as well as a span boundary objective to build robust multi-word representations. 
    \item We contribute finance-related benchmarks with 5 NLP tasks: financial sentiment analysis, news headline classification, named entity recognition, structure boundary detection, question answering. This results in a comprehensive suite of finance benchmarks, with licensing details in Table \ref{licensing}.
    \item We make all our models and code publicly available, for easier development and further research by the NLP and Finance community. Specifically, we contribute FLANG-BERT and FLANG-ELECTRA language models, and all the benchmarks in FLUE. 
\end{itemize}

\begin{table*}[]
\centering
\renewcommand{\arraystretch}{1.4}
\fontsize{20}{20}\selectfont
\resizebox{\textwidth}{!}{%
\begin{tabular}{|llllllllp{1.86cm}|}
\hline
Name & Task            & Source                       & \multicolumn{3}{c}{Dataset Size}  & Metric & License & Ethical Risks\\
 &             &                        & Train & Valid & Test  &  & &\\

\hline
FPB & Sentiment Classification  & \cite{finphrasebank} & 3488 & 388 & 969 & Accuracy & CC BY-SA 3.0 & Low\\
FiQA SA & Sentiment Analysis  & \cite{fiqa}{FiQA 2018} & 822 & 117 & 234 & MSE & Public & Low\\
Headline & News Headlines Classification & \cite{sinha2020impact} & 7,989 & 1,141 & 2,282 & Avg F-1 score & CC BY-SA 3.0 & Low\\
NER & Named Entity Recognition   & \cite{alvarado2015domain} & 932 & 232 & 302 & F-1 score &  CC BY-SA 3.0 & Low\\
FinSBD3 & Structure Boundary Detection & \cite{finsbd3}{FinWeb-2021} & 460 & 165 & 131 & F-1 score & CC BY-SA 3.0 & Low\\
FiQA QA & Question Answering & \cite{fiqa}{FiQA 2018} & 5676 & 631 & 333 & nDCG, MRR & Public & Low\\
\hline
\end{tabular}

}
\caption{Summary of benchmarks in FLUE. Dataset size denotes the number of samples in the benchmark. Metric denotes the evaluation metric used. Here MSE denotes Mean Squared Error, nDCG denotes Normalized Discounted Cumulative Gain and MRR denotes Mean Reciprocal Rank.}
\vspace*{-0.5em}
\label{licensing}
\end{table*}

\section{Related Work}
\paragraph{Pre-trained language models}
Language models pre-trained on unlabeled textual data, 
such as BERT \cite{bert}, ELMo \cite{elmo} and ROBERTA \cite{roberta}, have significantly improved the state-of-the-art in many natural language tasks. Newer models introduce  different training objectives: BART \cite{bart} uses denoising auto-encoder objective for sequence-to-sequence pre-training;  Span-BERT \cite{spanbert} uses a pre-training methodology that predicts spans of text; ELECTRA \cite{electra} uses token detection for training, where it corrupts some tokens using a generator network and predicts if the tokens are corrupted using a discriminator.

\paragraph{Masked Language Modeling}
Most language models use Masked Language Modeling (MLM) \cite{bert} as a training objective. It typically involves randomly masking a percentage of tokens in a text, and using surrounding text to predict the masked tokens. A variety of masking techniques have been used for domain-specific pre-training. While some works \cite{masking2, ernie} propose rule based masking strategies that work better than random masking, other works \cite{neuralmask} attempt to find optimal masking policy automatically using techniques such as reinforcement learning. 

\paragraph{Domain-specific Language Models}
While the models trained on general English language perform well, domain-specific pre-training can further increase the performance on a particular domain of text \cite{domain1, domain2}. For example, BioBERT \cite{biobert} on biomedical domain, ClinicalBERT \cite{clinicalbert} on clinical domain, SciBERT \cite{scibert} on scientific publications domain, etc. There have been some works on financial domain as well: previous works by \citet{orig_finbert, finbert} directly fine-tune BERT trained on financial corpus for sentiment analysis and question answering tasks respectively. FinBERT \cite{liu2020finbert} uses multi-task pre-training to improve performance. The previous works in financial domain rely on basic architectures/ training schemes and do not use finance-specific knowledge. 
Furthermore, FinBERT is pre-trained with the objective of optimizing performance for sentiment analysis, while we build a generalizable model performing well on a diverse set of tasks.  
We use and demonstrate that finance specific knowledge and vocabulary can further improve the performance of the model.
\paragraph{Finance Benchmarks}
\citet{glue} created General Language Understudy Evaluation(GLUE), a collection of benchmark tasks for training, evaluating, and analyzing language model designed for non-domain specific tasks. For financial domain, the benchmark suite isn't as exhaustive. \citet{finphrasebank} created Financial PhraseBank dataset for Sentiment analysis classification. \citet{FiQA10.1145/3184558.3192301} created two tasks in \cite{fiqa}{FiQA 2018}: Task-1 for Sentiment Analysis Regression and Task 2 dataset for Question Answering task in finance. Other datasets include gold news headline datase \cite{sinha2020impact}, financial NER \cite{alvarado2015domain} and Structure Boundary Detection \cite{finsbd3}. Recent financial language models \cite{orig_finbert, finbert} evaluate their efficacy only on sentiment analysis tasks. We use datasets from existing literature and create a set of heterogeneous benchmark tasks FLUE (Financial Language Understanding Evaluation) for better comprehensive evaluation.

\section{Benchmarks (FLUE) and Datasets}
\subsection{FLUE}
\label{subsec_FLUE}
We introduce Financial Language Understanding Evaluation (FLUE), a set of comprehensive benchmarks across 5 financial tasks. The statistics for FLUE are summarized in Table 1 along with the licensing details for public use. All FLUE benchmark datasets have low ethical risks and do not expose any sensitive information of any organization/ individual. Additionally, we have obtained approval for the authors of each dataset for this FLUE benchmark.

\subsubsection{Financial Sentiment Analysis} Serving as a fundamental task for textual analysis, this task received a lot of attention in finance domain \citep{loughran2011liability,garcia2013sentiment}. In our FLUE benchmark, we include both sentiment analysis tasks: regression and classification. For classification, we use Financial PhraseBank dataset \cite{finphrasebank} which provides the sentiment labels annotated by humans for financial news sequences. 
For regression, we use FiQA 2018 task-1 (Aspect-based financial sentiment analysis) dataset \cite{FiQA10.1145/3184558.3192301}, which  contains both headlines and microblogs. 

\subsubsection{News Headline Classification}
The financial phrases contain information on multiple dimensions other than the sentiment. Financial news headlines contain important time sensitive information on price changes. To explore our model on those dimensions, we use the Gold news headline dataset created by \citet{sinha2020impact}. The dataset is a collection of 11,412 news headlines, with 9 binary labels. 
 
\subsubsection{Named Entity Recognition} 
Name entity recognition (NER) is key task to analysing any financial text as it can be used along with the Knowledge Graphs to better understand interdependence of different financial entities linked through location, organisation and person. Given a text, NER can identify and classify tokens into specified categories such as person, organisation, location and miscellaneous. We use dataset released by \citet{alvarado2015domain} for NER task on financial domain text. 

\subsubsection{Structure Boundary Detection}
Boundary detection of different structure is fundamental challenge in processing text data. Here we employ the dataset shared in the task FinSBD-3 of \cite{finsbd3}{FinWeb-2021} workshop. The goal of the task is to find the boundaries of different components of text (sentences, lists and list items, including structure elements like footer, header, tables). We chose this dataset as it not only identifies boundaries of sentences but also identifies boundaries of other structural elements.  


\subsubsection{Question Answering} 
Question answering system which can answer the finance domain question is essential to any digital assistant. To evaluate our language model's ability on QA task we employ the dataset ("Opinion-based QA over financial data") released in \cite{fiqa} {FiQA 2018} open challenge Task 2 \cite{FiQA10.1145/3184558.3192301}.

\subsection{Pre-training Datasets}

For pre-training, we use a mix of general English language datasets and finance specific datasets. For English, we use BooksCorpus \cite{bookcorpus} (800M words) and English Wikipedia (2500M words). For the domain specific datasets, we use six publicly available datasets, they are: 1) SEC 10-K and 10-Q financial reports, 2) Earning Conference calls, 3) Analyst Reports, 4) Reuters Financial News, 5) Bloomberg Financial News, and 6) Investopedia. The details for these datasets are summarized in Table \ref{pretraining_datasets} and a brief description of each dataset is given in the Appendix Section \ref{app:pretraining_datasets}.

\begin{table*}[]
\centering
\begin{tabular}{|lllllll|}
\hline
Model  & FPB & FiQA SA & Headline & NER & FinSBD3 & FiQA QA  \\
Metric & Accuracy & MSE & Mean F-1 & F-1 & F-1 & nDCG  \\
\hline
BERT-base   &     0.856        &   0.073   &   0.967  & 0.79 & 0.95 &  0.46   \\
FinBERT \cite{finbert}      & 0.872          &   0.070   &  0.968   & 0.80 &   0.89    & 0.42  \\
FLANG-BERT(ours) &     \textbf{0.912}      &  \textbf{0.054} & \textbf{0.972}    & \textbf{0.83} &  \textbf{0.96}     & \textbf{0.51} \\
\hline
ELECTRA                           &    0.881         &    0.066  &  0.966   &     0.78    &         0.94              &       0.52            \\
FLANG-ELECTRA(ours)     &  \textbf{0.919}    &   \textbf{0.034}   &  \textbf{0.98}   &    \textbf{0.82}     &   \textbf{0.97}    & \textbf{0.55}   \\  
\hline
\end{tabular}

\caption{Summary of results of our models and baselines on benchmarks. FLANG (Financial Language Model) denotes our final model. Average of 3 seeds was used for each model and benchmark.}
\label{tb:results}
\vspace{-1em}
\end{table*}



\begin{table}[h]
\centering
\resizebox{0.40\textwidth}{!}{ 
\begin{tabular}{|lll|}
\hline
Model               &  MSE & R2 \\
\hline
SC-V \cite{yang2018financial_senti} & 0.080 & 0.40 \\
RCNN \cite{piao2018financial_senti} & 0.090 & 0.41 \\
\hline
BERT                  & 0.074      &  0.59  \\
FinBERT               & 0.070      &  0.57\\
FLANG-BERT            & \textbf{0.052}       &  \textbf{0.67}     \\
\hline
ELECTRA         & 0.046      & 0.72  \\
FLANG-ELECTRA      & \textbf{0.039}      & \textbf{0.77}     \\
\hline
\end{tabular}
}
\caption{Resuls on FiQA Sentiment Regression. }
\label{tb:FiQA_SA_res}
\end{table}



\begin{table}[]
\centering
\resizebox{0.35\textwidth}{!}{
\begin{tabular}{|lll|}
\hline
Model               &  \multicolumn{2}{l|}{F-1 Scores}  \\
\hline
Multi-token         &  No & Yes \\
\hline
CRFs            & \multicolumn{2}{c|}{{0.83}}  \\
\hline
BERT                  & 0.805    &   0.788  \\
FinBERT               & 0.795    &  0.800  \\
FLANG-BERT            &  \textbf{0.836}    &  \textbf{0.831}     \\
\hline
ELECTRA         & 0.797   & 0.777       \\
FLANG-ELECTRA  & \textbf{0.822} & \textbf{0.818} \\
\hline

\end{tabular}
}
\caption{Results on Named Entity Recognition. Yes: Set other tokens in word to same label. CRF result is taken from \cite{alvarado2015domain}, but they don't specify that whether they set other tokens in word to same label.}
\label{tb:NER_res}
\vspace*{-1.1em}
\end{table}

\section{Model}
\begin{table}[t]
\centering
\resizebox{0.44\textwidth}{!}{%
\begin{tabular}{|lll|}
\hline
Model               &  Accuracy & $\% \Delta MP $ \\
\hline
BERT                  &  85.6  &\\
FinBERT               &  87.2  & \\
FLANG-BERT       &   \textbf{91.2} &  31.25\\
\hline
ELECTRA         &  88.1  & 7.03  \\
\quad w/ AD      &  91.1    &   30.47\\
\quad w/ AD + PFV &  91.4  &32.81\\
\quad w/ AD + PFV + SBO      &  91.9  &36.71\\
\quad w/ AD + PFV + SBO + SCL      &  \textbf{92.1}  &\textbf{38.28}\\
\hline
\end{tabular}
}
\caption{Results on Financial Phrase Bank Sentiment Classification Dataset \cite{finphrasebank}. Accuracy is given as a percentage. Average of 3 seeds was used for all models. Marginal increase in performance is calculated for FLANG-ELECTRA with respect to FinBERT. FV means using Financial Vocabulary for masking, PFV means using both words and phrases in the financial dictionary for multi-stage masking in the pre-training task, SCL means the use of Supervised Contrastive Learning during the fine-tuning stage.}
\label{tb:FPB_SA_res}
\vspace*{-0.8em}
\end{table}

\begin{table*}[t]
\centering
\resizebox{2\columnwidth}{!}{%
\begin{tabular}{|l|l|ccc|cc|c|}
\hline
Category               & SVM & BERT & FinBERT & FLANG-BERT & ELECTRA & FLANG-ELECTRA & $\% \Delta MP $ \\
\hline
Price or Not           & \textbf{0.965} & 0.955 & 0.956 & 0.960 &  0.951   &   0.964    & 18.18  \\
Price Up               & 0.924          & 0.939 & {0.945} & 0.951 &  0.946   &  \textbf{0.964}   & 34.54  \\
Price Constant         & 0.715      & 0.980 & 0.978 & 0.981 &  0.977   &   \textbf{0.987}  & 40.90  \\
Price Down             & 0.932    & 0.950 & {0.958} & 0.965 &  0.959   &  \textbf{0.974}     & 38.09 \\
Past Price             & 0.965 & 0.947 & 0.952 & 0.955 &  0.943   &  \textbf{0.975}    & 47.91  \\
Future Price           & 0.732 & {0.987} & 0.985 & \textbf{0.988} &  0.984   &  \textbf{0.988}    & 20.00   \\    
Past News              & - & 0.950 & 0.951  & 0.952  &  0.945   &  \textbf{0.956}  & 10.20 \\
Future News            & - & 0.989 & 0.993  & 0.993  &  0.991   &  \textbf{0.994} & 14.28 \\  
Asset Comparison       & 0.994 & {0.998} & {0.998} & \textbf{0.999}   & 0.996 & 0.998 & 0  \\
\hline
Mean F-1 Score         & 0.890($\bar{7}$) & 0.967 & 0.968 & 0.973 & 0.966  & \textbf{0.978} & 31.25\\
\hline

\end{tabular}}
\caption{Results on News Headline Classification. SVM results are taken from \cite{sinha2020impact}. All  values are F1 scores. FLANG denotes our model. Average of 3 seeds was used for all models. FLANG-ELECTRA also uses Supervised Contrastive Learning while fine-tuning. Marginal increase in performance is calculated for FLANG-ELECTRA with respect to FinBERT.}
\label{tb:headline_res}
\vspace*{-0.8em}
\end{table*}

For \texttt{FLANG-BERT}, we add financial word and phrase masking, while for \texttt{FLANG-ELECTRA}, we also add a span boundary objective. The addition of financial word and phrasal masking is model agnostic and can be used for any model with a generator.

\subsection{Financial Word Masking}
Previous works \cite{liu2020finbert, finbert, orig_finbert} on financial language modeling use MLM objective for pre-training, which masks some tokens randomly and uses the prediction of those tokens as a training objective. However, there is empirical evidence \cite{ernie, neuralmask, masking2} that masking some words strategically which carry more information improves performance on downstream tasks.

Hence, we propose masking financial words preferentially. To this end, we use  Investopedia Financial Term Dictionary \cite{investopedia_dict} to create a comprehensive financial dictionary, which lists the commonly used technical terms in financial markets and literature. We expand our list by adding words/phrases from other financial vocabulary lists available online \cite{financial_dict_1, financial_dict_2, financial_dict_3}. 

Our dictionary contains more than 8200 words and phrases. For preferential masking, we mask the single word financial tokens with a probability 30\% and randomly mask other tokens with 70\% percent probability. Like original BERT pre-training scheme, we mask a cumulative total of 15\% of all tokens, such that the total number of tokens being masked in each round is same as the original BERT pre-training approach. Table \ref{tb:masking_perplexity} shows that masking financial terms with a 30\% probability gives the lowest perplexity score when pre-training either BERT and ELECTRA with additional vocabulary.

\subsection{Phrase Masking}

Many financial terms are phrases with multiple tokens. It has been shown \cite{ernie, spanbert} that masking phrases instead of words could leads to better learning of the phrase content. Building on that, we use phrase-based masking in the language model. We perform a two-phase training: in the first phase, we only use word masking to mask single tokens and train the language model; in the second phase, we add phrase masking.

For a financial term of token length $n$, we mask it with a probability of 30\%. We replace all tokens in a financial phrase with a single \texttt{[MASK]} token.
We add all the financial phrases in the model vocabulary and predict the phrase with the usual masked language modeling objective.
\subsection{Span Boundary Objective}
We add the Span Boundary Objective to the loss function along with the MLM loss in the pre-training stage,  in addition to the word and the phrasal level masking and the modified vocabulary. Our final loss has three parts: 
\paragraph{Masked Language Modeling Loss} is the Maximum Likelihood Loss of the ELECTRA generator (\textit{G}). 
We also modify the token masking to randomly mask contiguous spans from a geometric distribution of length L $\sim$ Geo(p), which is skewed towards smaller spans. We follow the results of \citet{spanbert} and set p = 0.2. 
$$L_{MLM}(x, \theta_G) = E(\sum_{i \in masks} - log(P_G (x_i|x_{masked}))$$
\paragraph{Discriminator loss} This loss term is the standard ELECTRA implementation. $L_{Disc}$ penalizes if the discriminator detects a token generated by the generator as \textit{replaced} when it is a \textit{non-corrupt} token or if the token generated by G is \textit{corrupt} and the discriminator detects it as \textit{original}.

\paragraph{Span Boundary Objective} This term penalizes the low probability of a token being generated given span boundaries (the the representations of tokens present before and after the masked contiguous span). The position of the left boundary token is $x_{start-1}$ and the position of the right boundary token is $x_{end+1}$. By looking at words before and after spans and then trying to generate the tokens in the span, this term helps the model to build multi-word  representations of financial terms that are not captured in our vocabulary. \\
    $$L_{SBO}(x, \theta_G)=E(\sum_{i \in masks} - log(P_G (x_i|y_i)))$$  $$\text{where } y_i = f(x_{start - 1}, x_{end+1}, pos_{i - start+1})$$
Here the function $f(c\dot)$ is the representation function for the $i^{th}$ token in the span and is defined by two feed forward layers:
$$y_i = \text{LayerNorm(GELU}(w_2* h1)$$
$$\text{where }h_1 = \text{LayerNorm(Gelu}(w_1*h_0))$$
$$\text{and } h_0 = [x_{start-1},x_{end+1},pos_i]$$
Our model is then pre-trained and optimized based on this combined loss function.
$$\text{Total Loss} = L_{MLM}(x, \theta_G) + \lambda_1 L_{SBO}(x, \theta_G)$$ $$ + \lambda_2 L_{Disc}(x, \theta_D)$$
\begin{table}[]
\centering
\resizebox{0.40\textwidth}{!}{  
\begin{tabular}{|lll|}
\hline
Model               &  \multicolumn{2}{l|}{F-1 Scores}  \\
\hline
Special tokens         &  No & Yes \\
\hline
BERT                  & 0.950    &   0.948  \\
FinBERT               & 0.872    &  0.890  \\
FLANG-BERT            &  0.964 & 0.958 \\
\hline
ELECTRA         & 0.938   & 0.968   \\
FLANG-ELECTRA  & 0.966* &  0.967*    \\
\hline
\end{tabular}}
\caption{Results on Structure Boundary Detection. *indicates the best model when the combined F1 score of both special tokens is considered. Yes and No are additional special tokens. Average of 3 seeds was used.
}
\label{tb:SBD_res}
\vspace*{-1em}
\end{table}
\subsection{Contrastive Loss for Fine-tuning}
While most language models are fine-tuned for supervised classification by using cross-entropy loss \cite{bert, roberta, gpt2}, we use additional supervised contrastive learning loss for fine-tuning for classification \cite{contrastiveloss}. This loss function captures the similarities between examples of the same class and contrasts them with the examples from other classes. Details about Supervised Contrastive Loss are given in Appendix Section \ref{app:constrastive_loss}.
Here, we only add this loss to the fine-tuning of Financial Phrasebank Dataset and the Headlines Dataset as shown in Tables \ref{tb:FPB_SA_res} and \ref{tb:headline_res}.

\begin{table}[]
\centering
\resizebox{0.42\textwidth}{!}{
\begin{tabular}{|llll|}
\hline
Model                 &  nDCG & MRR & Precision  \\
\hline
BERT                  & 0.46  &   0.42 & 0.35 \\
FinBERT               & 0.42  &  0.37  & 0.29 \\
FLANG-BERT            & 0.51  &  0.46  &  0.36    \\
SpanBERT + AD + FV + PFV            & \textbf{0.57}  &  \textbf{0.54}  &  \textbf{0.50}    \\
\hline
ELECTRA         & 0.52 & 0.49 & 0.43 \\
FLANG-ELECTRA     &   \textbf{0.55}  &  \textbf{0.51}  &  \textbf{0.45}   \\
\hline

\end{tabular}}
\caption{Results on Question Answering benchmark. Average of 3 seeds was used for all models.} 
\label{tb:QA_res}
\vspace*{-0.8em}
\end{table}

\section{Experiments}

\subsection{Experiment Setup}
\label{sec:Experiments_setup}
All experiments were conducted with PyTorch \cite{pytorch} on NVIDIA V100 GPUs. We initialized each model with their respective pre-trained version on the Huggingface's Transformers library~\cite{huggingface}. We further pre-trained each model for 4 more epochs on the training data. We used 2 epochs with only single token masking and the later 2 epochs for both word and phrase masking. Using this multi-stage setup gives the lowest model perplexity as shown in Table ~\ref{tb:two_step_masking}. 

We used ELECTRA-base pre-trained model as our base architecture. ELECTRA corrupts the input by replacing tokens with words sampled from a generator and trains a discriminator model that predicts whether each token in the corrupted input was replaced by a generator sample. This enables it to learn from all input tokens rather than just masked out tokens and is a good fit for our preferential masking approach.
We compare our results  the following models:
\begin{itemize}\setlength\itemsep{0em}
    \item BERT-base and ELECTRA-base: We use the BERT-base model \cite{bert} and the ELECTRA-base model \cite{electra} from Huggingface \cite{huggingface} and fine-tuned it directly for our tasks.
    \item finBERT \cite{finbert}: We used finBERT model and fine-tune on our tasks. 
    \item \textbf{FLANG-BERT (ours)} (Financial LANGuage Model based on BERT): For direct comparison with finBERT, we use our method to train a BERT-base model on our training corpus in a multi-stage manner (Table ~\ref{tb:two_step_masking}), masking single tokens from financial vocabulary in the first stage and then masking both words and phrases in the second stage.
    \item ELECTRA w/ AD (Additional Data): The ELECTRA base model pre-trained on our financial training corpus.
    \item ELECTRA w/ AD + FV (Financial Vocabulary): The ELECTRA Base model is pre-trained on our training corpus, while masking single tokens from financial vocabulary with a higher probability. 
    \item ELECTRA w/ AD + PFV (Phrase Financial Vocabulary). The ELECTRA Base model pre-trained on our training corpus in a multi-stage manner (Table~\ref{tb:two_step_masking}), masking only single-word tokens from financial vocabulary in the first stage and masking both words and phrases in the second stage.
    \item \textbf{FLANG-ELECTRA} (Financial LANGuage Model based on ELECTRA): ELECTRA w/ AD + PFV (Phrase Financial Vocabulary) + SBO (Span Boundary Objective). It is pre-trained on our training corpus in the described multi-stage manner with the span boundary and in-filling training objective. 
    \item ELECTRA w/ AD + PFV + SBO + SCL (Contrastive Loss): We use our final language model (FLANG-ELECTRA) but add a contrastive loss term to fine-tune on supervised classification tasks.
\end{itemize}

\begin{table*}[!t]
\resizebox{\textwidth}{!}{%
\begin{tabular}{|lp{1.5cm}p{1.5cm}p{1.5cm}p{1.5cm}p{1.5cm}p{1.5cm}|}
\hline
Model                    & FBP &  Headline & NER & FinSBD3 & FIQA SA & FIQA QA\\
\quad Metric                    & Accuracy &  Mean F-1 & F-1 & F-1 & MSE & nDCG\\
\hline
BERT & 0.856 &  0.967   & 0.79   & 0.949 & 0.073 & 0.46 \\
BERT + AD & 0.902 & 0.968   & 0.811   & 0.954 & 0.058 & 0.47\\
BERT + AD + FV + PFV (FLANG-BERT) & 0.912 & 0.972   & \textbf{0.834}   & 0.962  & 0.054 & 0.51\\
\hline
Distilbert & 0.844 & 0.963   & 0.776   & 0.934 & 0.075 & 0.45 \\
Distilbert + AD     & 0.898 & 0.965 & 0.806  & 0.944 & 0.064 &0.46 \\
Distilbert + AD + FV + PFV       & 0.901 & 0.965 & 0.812 & 0.958 & 0.057 & 0.49 \\
\hline
SpanBERT & 0.852 &  0.962   & 0.774   & 0.935 & 0.078 & 0.53\\
SpanBERT + AD & 0.901 & 0.962   & 0.789   & 0.951  & 0.063 & 0.55\\
SpanBERT + AD + FV + PFV & 0.904 & 0.969   & 0.792   & 0.959 & 0.056 & \textbf{0.57} \\
\hline
ELECTRA & 0.881 & 0.966   & 0.782   & 0.954 & 0.066 & 0.52
 \\
ELECTRA + AD  & 0.911 & 0.973   & 0.803   & 0.959 & 0.052 & 0.53\\
ELECTRA + AD + FV + PFV       & 0.914 & 0.977 & 0.825 & 0.962 &  0.038 & 0.55  \\
ELECTRA + AD + FV + PFV + SBO (FLANG-ELECTRA) & \textbf{0.919} & \textbf{0.978} &  0.816 & \textbf{0.967}  & \textbf{0.034} & 0.56
 \\
\hline
\end{tabular}%
}
\caption{Ablation Studies: Average of three seeds were used for each model and benchmark }
\label{tb:ablation}
\end{table*}

\subsection{Benchmark Results}

Summarized results on all benchmarks of our model and baselines are shown in Table~\ref{tb:results}.

\subsubsection{FPB Sentiment Classification}
The results of sentiment classification on Financial Phrase Bank sentiment dataset are shown in Table~\ref{tb:results}. From the accuracy numbers listed in the Table~\ref{tb:results}, it is evident that FLANG-BERT improves hugely on performance of FinBERT and our final language model(FLANG-ELECTRA) significantly outperforms all the baseline models on the sentiment classification task on the Financial Phrase Bank dataset, achieving state of the art results. Results in Table ~\ref{tb:FPB_SA_res} highlight the importance of each step in our experiment setup described in Section~\ref{sec:Experiments_setup}. As the previous state of art performance on this dataset is already in the higher 80s, we use an additional metric: marginal increase in performance over FinBERT ($\Delta MP$) to demonstrate our techniques. We calculate ($\Delta MP$) as given in equation \ref{eq:1}:
\begin{equation}\label{eq:1}
    \Delta MP = \frac{Metric_{Model} - Metric_{FinBERT}}{1-Metric_{FinBERT}}
\end{equation}
where the Metric is Accuracy for the Financial Phrasebank Dataset and is F1 score for News Headlines Dataset.

\subsubsection{FiQA Sentiment Regression}
The results of sentiment regression analysis on the \citet{fiqa} dataset are shown in Table~\ref{tb:FiQA_SA_res}. 
Evaluation of models is done on two regression evaluation measures Mean Squared Error (MSE) and R Square (R2). Our transformer based architectures outperform conventional techniques like SC-V and RCNN. FLANG-BERT model achieves significant improvement on both BERT and finBERT and FLANG-ELECTRA outperforms all models and achieves state of art result for the sentiment regression analysis task on the FIQA dataset.

\subsubsection{News Headline Classification}
The results of news headline classification for 9 binary classification tasks on Gold headline dataset are shown in Table~\ref{tb:headline_res}. All the deep learning based language models perform much better than Support Vector Machines. Our ELECTRA-based language model (FLANG-ELECTRA) achieves the highest mean F-1 score compared to other language models. FLANG-BERT performs better than BERT, which again highlights the importance of our setup.

\subsubsection{Named Entity Recognition}
The results of NER on financial NER dataset provided by \cite{alvarado2015domain} are shown in  Table~\ref{tb:NER_res}. The margin of improvement is more muted in this benchmark. Our models outperform the baselines in a multi-token setting. The multi-token setting refers to all tokens in a word being set to the same label when a word is split into multiple tokens, instead of only labeling the first token and ignoring the rest. Our hypothesis is that when the task doesn't require domain specific knowledge, like NER, pre-training language model on domain specific data does not help. 
\begin{table}[]
\centering
\resizebox{0.46\textwidth}{!}{%
\begin{tabular}{|p{3cm}p{1cm}p{1cm}p{1cm}p{1cm}|}
\hline
\textit{Model Perplexities}&  \multicolumn{2}{c}{BERT} & \multicolumn{2}{c|}{ELECTRA}\\
\hline
\% of Financial Terms Masked &  FV & PFV &  FV  & PFV\\
\hline
\quad 10 & 23.02 & 22.88 & 19.10 & 18.96\\
\quad 20 & 21.45 & 21.30 & 18.44 & 18.42\\
\quad 30 & \textbf{20.29} & \textbf{19.53} & \textbf{17.87} & \textbf{17.52}\\
\quad 40 & 20.80 & 20.11 & 18.67 & 17.98\\
 \hline
\end{tabular}
}
\caption{ Model Perplexities when different percentages of Financial terms are masked. FV means using Financial Vocabulary for masking, PFV means using both words and phrases in the financial dictionary for multi-stage masking in the pre-training task.}
\label{tb:masking_perplexity}
\end{table}
\subsubsection{Structure Boundary Detection}
The results of structure boundary detection task on FinSBD3 dataset from \cite{finsbd3}{FinWeb-2021} are shown in Table~\ref{tb:SBD_res}. In this table, note that the "Special Tokens" setting refers to adding special tokens that are commonly used by pre-trained transformers such as [CLS] to the input. 
Our models perform similarly or slightly better to baseline architectures. This could be because SBD, like NER, relies more on language cues rather than finance keywords for inference and further gives evidence to the hypothesis that when the task doesn't require domain specific knowledge, one should not get improvement by pre-training a language model on domain specific data. However, our model still performs significantly better than FinBERT.

\subsubsection{Question Answering}
On Question-Answering, our models outperform the previous works, as shown in Table~\ref{tb:QA_res}. For evaluation, we compare the following metrics \cite{ndcg}: Precision, nDCG---A higher value means that more relevant documents are retrieved first, and MRR---A higher value means that the first relevant item is retrieved earlier. 
FLANG-BERT, FLANG-ELECTRA outperform other models on all metrics by a huge margin, but do not outperform SpanBERT pre-trained with Additional Data.

\subsection{Ablation Studies}
\begin{table}[]
\centering
\resizebox{0.36\textwidth}{!}{%
\begin{tabular}{|cccc|}
\hline
 \multicolumn{2}{|c}{Number of Epochs} & \multicolumn{2}{c|}{Model Perplexity}\\
\hline
 FV &  FV + PFV & BERT & ELECTRA \\
\hline
4 &  0 & 20.29 & 17.87 \\
3 & 1 & 20.11 & 17.82\\
2 &  2 & 19.53 & 17.52\\
1 &  3 & 20.13 & 17.80\\
0 &  4 & 20.05 & 17.69\\
 \hline

\end{tabular}
}
\caption{ Model Perplexities when using multi-stage financial term masking for pre-training. FV means using Financial Vocabulary for masking, PFV means using both words and phrases in the financial dictionary for multi-stage masking.}
\label{tb:two_step_masking}
\end{table}

We conduct multiple ablation studies to understand the individual impact of our techniques on performance. Our studies in Table~\ref{tb:masking_perplexity} show that preferentially masking 30\% of the financial tokens gives the least perplexity for each model. Furthermore, we find that using single-word financial terminologies in the first two pre-training epochs and multi-word terminologies in the next two gives the lowest perplexity score (Table~\ref{tb:two_step_masking}). Table~\ref{tb:ablation} shows that the use of additional data and domain specific preferential masking give substantial increase in performance for our FLUE tasks. Addition of the Span Boundary Objective on the ELECTRA generator gives the best performing model when compared to other similar encoder based architectures like SpanBERT, DistilBERT and BERT. In Table~\ref{tb:perplexity}, we also show that pre-training models using our methodology gives the lowest perplexity scores when compared to prior baselines. The details for the studies can be found in Table~\ref{tb:ablation} and Appendix Section~\ref{app:ablation}.

\begin{table}[]
\centering
\resizebox{0.36\textwidth}{!}{%
\begin{tabular}{|lll|}
\hline
Model & Perplexity & Size\\
\hline
BERT-base   &   23.66     & 110M \\
FinBERT       &   21.11   &  110M\\
FLANG-BERT &   19.53    &  110M\\
Electra  & 20.10 & 110M \\
\quad w/ AD & 19.20 & 110M \\
\quad w/ AD + FV & 17.87   & 110M  \\
\quad w/ AD + PFV &  17.52  & 110M\\
\quad w/ AD + PFV + SBO &  17.34  & 110M\\
\hline
\end{tabular}
}
\caption{Comparison of perplexity of our model and baselines. The model size is given in terms of number of parameters, and perplexity is averaged over all sentences in the validation dataset. Average of 3 runs was used for perplexity numbers. Here AD means Additional financial data, FV means using Financial Vocabulary for masking, PFV means using both words and phrases in the financial dictionary for multi-stage masking, and SBO means using the span boundary objective in the pre-training task.}

\label{tb:perplexity}
\end{table}

\subsection{Discussion}
In conclusion, both FLANG-ELECTRA and FLANG-BERT outperform the base architectures (ELECTRA and BERT, respectively). FLANG-BERT also outperforms FinBERT on all the benchmarks, with the same number of parameters. Additionally, on relatively domain-agnostic tasks such as Named Entity Recognition, the improvements are muted. The performance is hugely improved in tasks which utilize finance specific language, such as sentiment analysis, sentence classification and question answering.
Overall, the dramatic improvement in most benchmarks 
suggests that our technique yields state-of-the-art financial language models. We also note that our vocabulary based preferential masking training methodology is both architecture and domain independent and can be generalized to other language models and domains.
\section{Conclusion}
We contribute two language models in the finance domain, which use domain-specific word and phrase masking as a pre-training objective. Additionally, we contribute a comprehensive suite of benchmarks in finance domain across 5 natural language tasks, including new benchmarks using public sources. Our language model outperforms previous language models on all the benchmarks. We will release our models, code and benchmark data on acceptance. We also note that our method is not specific to finance and can be used for any domain-specific language model training.
\section*{Acknowledgements}
We would like to thank the anonymous reviewers
for their comments. We appreciate the generous support of Azure credits from Microsoft made available for this research via the Georgia institute of Technology Cloud Hub. This work is supported in part by the J.P. Morgan AI Faculty Research Award. Any opinions, findings, and conclusions in this paper are those of the authors only and do not necessarily reflect the views of the sponsors.

\section*{Ethics Statement}
We give full credit to the respective authors of each dataset included in our FLUE benchmark and have obtained their permissions for the inclusion of each dataset in FLUE. All FLUE benchmark datasets have low ethical risks and do not expose any sensitive or personal identifiable information. We also obtain explicit permissions to use the datasets given in section \ref{pretraining_datasets} for pre-training of the FLANG models from the respective sources. 

We understand that training large language models has big carbon-footprint and we have tried to minimize the number of full-scale pre-training runs. The addition of preferential masking and the span boundary objective have minimal computation overhead when compared to pre-training traditional BERT/ELECTRA. 
We hope that future models work towards lower carbon footprint to reduce the environment costs of pre-training for more sustainable and ethical AI.

\section*{Limitations}
Some limitations to our work are: 1) We have not included abstractive generation or summarization tasks in the FLUE benchmark, due to a lack of large, annotated datasets. Future work can be directed towards summarization efforts for the financial domain. 2) We do not include social media data like twitter and reddit in our pre-training step, despite the heavy impact of social media on some financial markets like crypto currencies. This is because of the informal usage of textual data which impedes the formal and syntactical correctness of most financial documents. 3) The models are trained and tested on English tasks and may not perform well on non-English text. The limited availability of non-English domain specific vocabulary makes building multi-lingual FLANG models difficult. 4) While the methodologies presented in this paper can work well for any similarly structured domain like clinical data, it is often difficult to obtain a vocabulary term lists and dictionaries for certain domains. 5) We limit ourselves to using encoder based architectures due to the nature of the popular financial domain specific tasks. Future works can explore the use of other models like GPT3 and T5 for the domain.

\bibliographystyle{acl_natbib}
\bibliography{anthology,custom}
\section{Appendix}
\begin{table*}[]
\centering
\begin{tabular}{|lllll|}
\hline
Name & Source & Size  & Time Period & \%age sampled\\
\hline
10-K  &   SEC EDGAR  & 13660 & 1993-2020           & 8 \\
10-Q  &   SEC EDGAR  & 36402 & 1993-2020          &  5 \\
Earning Call Transcripts    &   {SeekingAlpha}   & 151359 & 2007-2019 & 1.5  \\
Financial News &  {Reuters TRC2 Corpus} & 106521 & 2007 & 10 \\
Financial News  &  {Bloomberg Corpus} & 387220 & 2009 & 5 \\
Analyst Reports & LexisNexis & 201 & 2017-2020 & 100 \\
Investopedia Articles & {Investopedia} & 638 & NA & 100 \\
\hline
\end{tabular}
\caption{Summary of financial datasets used for pre-training. Model size denotes the number of samples in the dataset. \%age sampled denotes the percentage of each dataset we sampled in a single training epoch.}
\label{pretraining_datasets}
\end{table*}


\subsection{Pre-training datasets}
\label{app:pretraining_datasets}

Table \ref{pretraining_datasets} summarizes the financial datasets used for pre-training. It also presents the percentage of each dataset sampled in one training epoch. A brief description of each dataset used for pre-training is given below:
\subsubsection{SEC Financial Reports} Most U.S. public firms are required by the U.S. Securities and Exchange Commission (SEC) to submit annual report (10-K) and quarterly report (10-Q), to provide detailed information about the firm's business, risk factors, and financial performance. 10-K and 10-Q filings were analyzed in \citep{li2010information,loughran2011liability, buehlmaier2018financial, chava2016december}. We download the 10-K and 10-Q filings from SEC EDGAR during 1993--2020. 

\subsubsection{Earnings Conference Calls} The earnings conference calls are held by public companies to covey critical corporate information to the investors and analysts \citep{bushee2003open, bowen2002conference}. SeekingAlpha, as a crowd-sourced website in the United States, provides investing information for a large number of public companies and publishes textual transcripts of many earnings conference calls. \citet{bochkay2020hyperbole} use the earnings conference call transcripts to analyze the stock market response to the language extremity. \cite{chava2019buzzwords} use BERT to construct emerging technology related discussions in earnings calls and evaluate whether it is just hype. \cite{chava2020ESG} employ RoBERTa to extract environmental related discussion in earnings calls and analyze whether managers walk their talk. We collect 151,359 earnings call transcripts from SeekingAlpha from Jan. 2000 to Jul. 2019. \cite{chava2022measuring} use BERT to construct a text-based firm-level inflation exposure measure on earning call transcripts.

\subsubsection{Analyst Reports} Security analysts generate reports related to a firms' future performance after collecting and analyzing the relevant information. Most analyst reports contains earnings forecast, stock recommendation, and stock price target \citep{asquith2005information}. We collect around 201 analyst reports on public firms from LexisNexis. This corpus contains the language the analysts use to disseminate the new information and their interpretation of previous released information to the investors.

\subsubsection{Reuters Financial News} Financial news corpus is helpful in analyzing the language used in business society. The Thomson Reuters Text Research Collection (TRC2) contains over 1.8M financial news stories during 2008--2009, which is deployed in prior literature \citep{orig_finbert}. We use 10\% of this corpus to pre-train our model.


\subsubsection{Bloomberg Financial News} 
Bloomberg disseminates business and market news to the market investors. We obtain the publicly available Bloomberg news articles provided by \citet{BloombergReutersDataset2015}, which is used in \citet{ding2014using} to predict the return of Standard \& Poor’s 500 stock (S\&P 500) index.

\subsubsection{Investopedia}
Investopedia is a financial website which serves as a comprehensive financial dictionary and provides definition and explanation for financial terminologies used in business world. We download the 638 articles for the financial concepts, and use them to pre-train our model. These articles not only provide definitions of financial terms, but also show how they are interrelated to each other.

\subsection{Ablation Studies}

\label{app:ablation}

\subsubsection{Preferential Masking with Financial Vocabulary}
For the first study, we try different configurations while preferentially masking financial terms in the pre-training. Table ~\ref{tb:masking_perplexity} shows the impact of masking different percentages of Financial Terms on the model perplexity. The perplexities are calculated while keeping the total percentage of masked tokens for all vocabulary at 15 percent. Table ~\ref{tb:masking_perplexity} shows that masking 30 percent of financial terms gives the least perplexity on the validation set. We also experiment with the multi-stage masking, where in the first stage (first n epochs) we use only the single-word financial tokens and in the second stage (next m epochs) we use both: word and phrasal financial vocabulary masking. Table ~\ref{tb:two_step_masking} shows that masking single-word financial vocabulary in the first 2 epochs and masking all financial terms has the lowest perplexity score.

\subsubsection{Perplexity on Validation Set}

For the second study, we compute perplexity of the language model on the validation set after pre-training. We report the perplexity scores in Table~\ref{tb:perplexity}. We notice that FLANG-BERT significantly lowers the perplexity on validation set, relative to BERT and FinBERT \cite{orig_finbert}. Despite all models having the same number of parameters, ELECTRA based models show lower perplexity scores. For ablation study, we keep ELECTRA architecture fixed and notice that pre-training with financial data along with general English data lowers perplexity compared to base ELECTRA. Further reduction is seen when using our token masking approach with financial keywords, suggesting that domain specific masking is helpful for domain specific language models. Pre-training with phrase based masking with the span boundary objective in the generator stage results in the best performance, validating the performance of our technique.

\subsubsection{FPB Sentiment Classification}
\label{app:sentiment}
For the third study, we fine-tune the models for sentiment analysis on the Financial PhraseBank Dataset \cite{finphrasebank} and report the accuracy in Table~\ref{tb:FPB_SA_res}. We perform a detailed ablation study on ELECTRA architectures with our various techniques. The results suggest that pre-training on financial data improves accuracy from 88.1\% to 91.1\%, and using a financial vocabulary for token masking further improves the performance to 91.4\%. Span boundary objective is even more effective, improving accuracy to $ >91.5\% $. Using contrastive learning for fine-tuning further enables an accuracy of 92.1\%, which is significantly higher than previous works.

\subsection{Supervised Contrastive Loss}
\label{app:constrastive_loss}
Language models are usually fine-tuned \cite{bert, roberta, gpt2} for supervised classification tasks by using cross entropy loss $L_{CE}$:
\begin{equation}
    L_{CE} = -\frac{1}{N}\sum_{i=1}^{N} \sum_{i=1}^{C} y_{i,c} \; log \; \hat{y}_{i,c}
\end{equation}
where $N$ is the number of samples, $C$ is the number of classes, $(x_i, y_i)$ are the sentence and label pairs for sample $i$ and $\hat{y}_{i,c}$ is the model output for probability of sample $i$ having class $c$.

\citet{contrastiveloss} showed that using an additional supervised contrastive learning loss $L_{SCL}$ for fine-tuning pre-trained language models improves performance. The loss is meant to capture the similarities between examples of the same class and contrast them with the examples from other classes:

\begin{equation}
\begin{split}
    L_{SCL}& = \sum_{i=1}^{N} -\frac{1}{N_{y_i}-1} \sum_{j=1}^{N} \mathbb{1}_{i \neq j} \mathbb{1}_{y_i = y_j} ( \\
    & log\frac{exp(\phi(x_i) \dot \phi(x_j))}{\sum_{k=1}^{N}\mathbb{1}_{i\neq k} exp(\phi(x_i) \dot \phi(x_k)))})
\end{split}
\end{equation}
where $N_{c}$ is the number of samples of class $c$.

Overall loss is given by:
\begin{equation}
    L = \lambda L_{CE} + (1 - \lambda) L_{SCL}
\end{equation}
where $\lambda$ is a variable for weighing the two losses.
\end{document}